\documentclass{article}
\usepackage{iclr2027_conference,times}

\usepackage{amsmath,amsfonts,bm}

\newcommand{\captiona}{{\em (a)}}
\newcommand{\captionb}{{\em (b)}}

\def\eqref#1{equation~\ref{#1}}
\def\Eqref#1{Equation~\ref{#1}}

\def\1{\bm{1}}

\DeclareMathAlphabet{\mathsfit}{\encodingdefault}{\sfdefault}{m}{sl}
\SetMathAlphabet{\mathsfit}{bold}{\encodingdefault}{\sfdefault}{bx}{n}

\usepackage{hyperref}
\usepackage{url}
\usepackage{graphicx}
\usepackage{booktabs}
\usepackage{algorithm}
\usepackage{algorithmic}
\usepackage{float}

\title{InterQ-RL: Interaction-Aware Correction of Execution Mismatch in Quantized Agentic RL}

\author{\makebox[\dimexpr\textwidth-2\tabcolsep\relax][c]{Yingbo Hao$^{1,2}$, Song Liu$^{2}$, Guanghua Yu$^{2}$, Zhaojing Zhou$^{2}$, Jianchen Zhu$^{2}$, Yi Zou$^{1}$} \\
\makebox[\dimexpr\textwidth-2\tabcolsep\relax][c]{$^{1}$South China University of Technology \quad $^{2}$Tencent Hunyuan}
}

\iclrfinalcopy 
\begin{document}

\maketitle
\lhead{}
\renewcommand{\headrulewidth}{0pt}

\begin{abstract}
Multi-turn agent reinforcement learning (RL) relies on long closed-loop rollouts, making generation
and KV-cache costs a major bottleneck.  Quantized rollout reduces these
costs, but its low-precision executor also serves as the behavior policy and
can induce execution mismatch with the learner.  Existing quantized-RL
methods are designed primarily for single-turn generation.  They align the
learner forward with quantized rollout or correct the current-token
rollout--learner ratio, without modeling the policy history that produces
each agent state.
Even after quantization alignment, residual rollout--learner mismatch accumulates
along the state-reaching policy history, so a current action may look locally
aligned despite a mismatched history.  We introduce
\textbf{InterQ-RL} (Interaction-aware Quantized Reinforcement Learning) to
correct this post-alignment mismatch.  It applies bounded execution-prefix
correction over the state-reaching policy history, allocates correction
strength across action segments by effective support under a fixed displacement
budget, and requires no additional model forward pass.
We evaluate dense and mixture-of-experts (MoE) backbones from 4B to 30B on ALFWorld, WebShop, and
ScienceWorld, with all quantized arms sharing the same rollout with 8-bit
weights, activations, and KV cache (W8A8-KV8).  Across four
environment--backbone settings, InterQ-RL achieves the strongest quantized
endpoint result, matching BF16 to within 0.7 points and
exceeding it in two settings.  It matches or exceeds the learning speed of
alignment alone while avoiding its late performance regression.
Code is available at \url{https://github.com/Damon-Salvetore/InterQ-RL}.
\end{abstract}

\section{Introduction}

On-policy PPO/GRPO optimizes language-agent outcomes using fresh trajectories
from the current policy at every update
\citep{guo_2025_deepseekr1,shao_2024_deepseekmath}.  These trajectories can
span multiple turns of search, environment interaction, and tool use
\citep{yao_2023_react,cheng_2026_agentr1,feng_2025_gigpo}.  To scale this
data-generation loop, modern systems decouple rollout generation from learning
\citep{sheng_2025_hybridflow,kwon_2023_vllm,hu_2024_openrlhf,fu_2025_areal}.
As interactions grow longer,
autoregressive generation and the expanding KV cache become major compute and
memory bottlenecks on the rollout side.  W8A8 reduces linear-operator cost,
and KV8 reduces the cache footprint of long-context rollout
\citep{micikevicius_2022_fp8,qiu_2026_fp8rl}.  Yet the low-precision rollout
executor is also the behavior policy.  When the executor and learner evaluate
the same checkpoint along different numerical paths, they can define
different action distributions.  We call this
\emph{execution mismatch}, which violates the behavior--old-policy
consistency assumed by PPO/GRPO
\citep{schulman_2017_ppo,shao_2024_deepseekmath}.

Prior quantized-RL work addresses training--inference mismatch
\citep{zhong_2026_vexact} in two ways: alignment matches the learner forward
to quantized rollout
\citep{gu_2026_qarl,xi_2026_jetrl}, whereas
correction clips or reweights the rollout--learner ratio at the
current token
\citep{yao_2025_rollout_mismatch,zhou_2026_ais,li_2026_qurl,qiu_2026_fp8rl}.
Both lines focus on single-turn generation.  They do not address residual
rollout--learner mismatch that remains after alignment and accumulates over
the policy history of multi-turn agents.
In a closed-loop agent, earlier actions determine later states.  Along the
same realized trajectory, shared environment-transition factors cancel, but
the ratios of policy actions remain
\citep{sutton_2018_rl,precup_2000_offpolicy}.  Environment feedback changes
later contexts but contributes no policy-likelihood factor.  We call this
\emph{interaction-structured execution mismatch}.

\begin{figure}[t]
  \centering
  \begin{minipage}[t]{0.5125\linewidth}
    \centering
    \includegraphics[width=\linewidth]{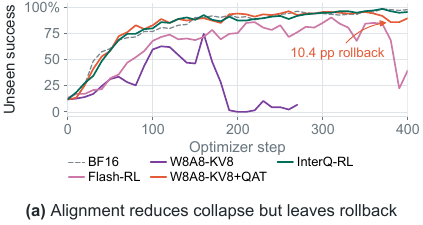}
  \end{minipage}\hfill
  \begin{minipage}[t]{0.4655\linewidth}
    \centering
    \includegraphics[width=\linewidth]{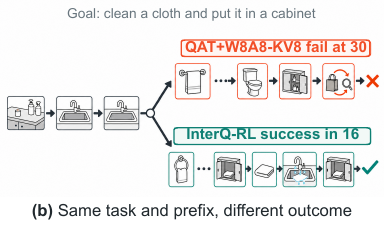}
  \end{minipage}
  \caption{\captiona\ ALFWorld
  \texttt{valid\_unseen} success, evaluated every 10 steps. W8A8-KV8 is the
  unaligned run and collapses; W8A8-KV8+QAT is aligned but drops 10.4 points
  from step 260 to 380 and only partially recovers by step 400. Flash-RL
  destabilizes late, while InterQ-RL remains
  stable and approaches BF16 at step 400.
  \captionb\ For one step-400 task, the two arms diverge after sharing their
  first three actions: W8A8-KV8+QAT exhausts the 30-action budget, whereas
  InterQ-RL finishes in 16. Across all 134 tasks, $10.45\%$ of
  W8A8-KV8+QAT episodes and $4.48\%$ of InterQ-RL episodes fail after
  exhausting the action budget.}
  \label{fig:agentic-motivation}
\end{figure}

We therefore propose InterQ-RL (Interaction-aware Quantized
Reinforcement Learning).  The method first uses quantization-aware training
(QAT) to align the learner forward with rollout quantization.  It then forms a bounded
execution-prefix correction by accumulating rollout--learner log-ratios along
the state-reaching policy history.  Each policy-generated action is treated
as one allocation segment.
For each segment, InterQ-RL computes an effective-support score from the
prefix weights and uses it to allocate correction under a fixed
squared-displacement budget.  The procedure reuses existing log probabilities
and requires no additional model forward pass.
Figure~\ref{fig:agentic-motivation} motivates correction beyond alignment:
the aligned baseline remains trainable but regresses late and exhausts the
action budget more often than InterQ-RL.

We evaluate InterQ-RL on three agentic environments and 4B--30B dense and MoE
Qwen3 backbones \citep{yang_2025_qwen3} under a common W8A8-KV8 rollout.
It outperforms QAT by 6.0 and 2.8 points on ALFWorld success and WebShop
score \citep{yao_2022_webshop} while matching BF16 within 0.7 points on
ALFWorld success, 0.3 on WebShop score, and 0.5 on WebShop success.  On both ScienceWorld
\citep{wang_2022_scienceworld} backbones, it attains the highest macro and
solved endpoints.  Its normalized area under the learning curve (AULC) is
within 0.33 of QAT on ALFWorld and highest among quantized arms in the
other three settings.

This paper makes the following contributions.

\begin{enumerate}
\setlength{\topsep}{3pt}
\setlength{\partopsep}{0pt}
\setlength{\itemsep}{1pt}
\setlength{\parsep}{0pt}
\item We characterize \emph{interaction-structured execution mismatch} in
quantized agentic RL: W/A+KV alignment leaves residual rollout--learner
mismatch that compounds along the state-reaching policy history.
\item We propose InterQ-RL, which applies bounded execution-prefix correction
over the state-reaching policy history, allocates correction strength across
action segments by effective support under a fixed displacement budget, and
requires no additional model forward pass.
\item Using a common W8A8-KV8 rollout, we evaluate three agentic environments
and 4B--30B dense and MoE backbones.  Across all four settings, InterQ-RL
achieves the strongest quantized endpoint result; its AULC is within 0.33 of
QAT on ALFWorld and highest among quantized arms in the other three
settings.
\end{enumerate}

\section{Preliminaries}
\label{sec:prelim}

\subsection{Policy Optimization with Quantized Rollouts}
\label{sec:prelim-po}

Let $q_{\mathrm{old}}$ be the learner policy at the rollout checkpoint and
$q_\theta$ the current policy.  In quantized rollout
\citep{li_2026_qurl,qiu_2026_fp8rl,zhou_2026_ais}, a token $x_t$ in
context $h_t$ is sampled from the behavior policy $r_{\mathrm{exec}}$
realized by the low-precision executor.  Even at a shared
checkpoint, $r_{\mathrm{exec}}$ and $q_{\mathrm{old}}$ may differ, so the
current-policy-to-behavior likelihood ratio factors as
\begin{equation}
\label{eq:two-ratios}
\frac{q_\theta(x_t\mid h_t)}{r_{\mathrm{exec}}(x_t\mid h_t)}
=
\underbrace{\frac{q_\theta(x_t\mid h_t)}{q_{\mathrm{old}}(x_t\mid h_t)}}_{\text{optimization ratio}}
\cdot
\underbrace{\frac{q_{\mathrm{old}}(x_t\mid h_t)}{r_{\mathrm{exec}}(x_t\mid h_t)}}_{\text{rollout--learner ratio}}.
\end{equation}
The optimization ratio is clipped by the PPO/GRPO objective
\citep{schulman_2017_ppo,shao_2024_deepseekmath}; the rollout--learner ratio
already differs from one before any parameter update, because the executor
evaluates the same weights in low precision.

QAT simulates W8A8-KV8 in the learner forward while retaining high-precision master
parameters, gradients, and optimizer states
\citep{liu_2024_llmqat,gu_2026_qarl}.

\subsection{Agentic Trajectories}
\label{sec:prelim-traj}

A language agent alternates policy actions with tool or environment feedback
\citep{yao_2023_react,cheng_2026_agentr1}.  For task $z$, an $S$-step
trajectory is
\begin{equation}
\tau=(z,o_0,a_1,o_1,\ldots,a_S,o_S),
\qquad R=R(\tau),
\end{equation}
where $o_0$ is the initial observation, $a_s$ a policy-generated action segment,
$o_s$ the subsequent observation, and $R$ the trajectory-level reward.
After flattening into tokens, set $m_t=1$ at policy-generated positions and
$m_t=0$ at observations and other non-policy positions.  Observations enter
neither the policy likelihood nor the policy-gradient loss, but remain in the
context $h_t$ of later actions.

\subsection{Trajectory-Prefix Likelihood Ratios}
\label{sec:prelim-prefix}

Let target policy $\pi$ and behavior policy $\mu$ share the task distribution,
environment transition kernel, and interaction protocol, and induce
trajectory distributions $P_\pi$ and $P_\mu$.  If $P_\pi\ll P_\mu$, the
likelihood ratio of a realized behavior prefix is
\begin{equation}
\label{eq:prefix-identity}
\frac{P_\pi(\tau_{\le t})}{P_\mu(\tau_{\le t})}
=
\prod_{\substack{u\le t\\ m_u=1}}
\frac{\pi(x_u\mid h_u)}{\mu(x_u\mid h_u)} .
\end{equation}
This is the standard trajectory change-of-measure identity
\citep{sutton_2018_rl,precup_2000_offpolicy}.  The policies are evaluated on
the same realized history and share the environment kernel, so initial-state
and transition factors cancel.  The product therefore retains only policy
positions from the realized prefix, while feedback remains action-conditioned
and enters later contexts.

\Eqref{eq:prefix-identity} is the exact prefix likelihood-ratio identity; it
determines the policy-history scope of the bounded training surrogate in
\S\ref{sec:method}.

\section{InterQ-RL: Aligning Execution and Correcting Interaction History}
\label{sec:method}

\begin{figure}[t]
  \centering
  \includegraphics[width=\linewidth]{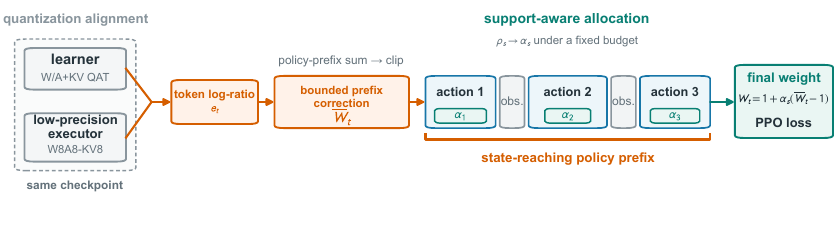}
  \caption{InterQ-RL aligns quantization, corrects the state-reaching policy
  history, and allocates correction strength by effective support under a
  fixed displacement budget. Grey: the learner forward matches rollout W/A and KV quantization at
  a shared checkpoint. Orange: per-token rollout--learner log-ratios form the
  prefix correction; feedback changes later contexts but adds no policy-likelihood term.
  Green: segment-level effective support $\rho_s$
  determines retention $\alpha_s$ under this budget.}
  \label{fig:method-overview}
\end{figure}

InterQ-RL first aligns the quantization semantics of learner and rollout, then
constructs a bounded correction surrogate by detached reweighting of the base
PPO/GRPO loss.
\emph{Scope} determines which policy history leading to the current state
enters the surrogate; \emph{allocation} determines how fixed correction
strength is distributed across action segments.  InterQ-RL addresses these
choices with execution-prefix correction and support-aware allocation,
respectively (Figure~\ref{fig:method-overview}).

\subsection{Aligning the Policy that Generated the Data}
\label{sec:method-align}

The first stage reproduces the rollout quantization semantics in the learner
forward.  Rollout uses blockwise W8A8 linear operators and stores K/V in FP8
E4M3 with per-token-per-head scaling.  Standard QAT covers the W/A path.  For
the KV path we match the learner quantize--dequantize rule to the rollout
cache-write kernel, at the same points in the layer.
Appendix~A gives the full specification.  The two forward
paths thereby share quantization semantics, while their conditional
distributions may still differ at a shared checkpoint.

For a rollout token $x_t$ in context $h_t$, we quantify this residual
mismatch by
$\delta_t:=\log q_{\mathrm{old}}(x_t\mid h_t)
-\log r_{\mathrm{exec}}(x_t\mid h_t)$.  On executor samples, we report the
standard nonnegative K3 estimator for
$D_{\mathrm{KL}}(r_{\mathrm{exec}}\Vert q_{\mathrm{old}})$
\citep{schulman_2020_klapprox,shao_2024_deepseekmath}.  We use K3 as a
diagnostic; Appendix~A gives its formula and expectation.
Over the final 20 steps in Figure~\ref{fig:qat-gap}, median K3 falls from 1.03
for W8A8-KV8 to 0.029 for W8A8-KV8+QAT, a reduction of roughly one and a half
orders of magnitude.  The aligned value nevertheless remains about
20$\times$ the BF16 value of 0.0014.  Alignment therefore removes most of the token-local
mismatch; InterQ-RL uses the remainder to construct its history correction.

\subsection{Correcting the History that Produced the Current State}
\label{sec:method-prefix}

After alignment, token-local correction still compares only the conditional
probabilities of the current token $x_t$ in the realized context $h_t$; it
omits how the policies differ in reaching that context.

Take the target and behavior policies of \S\ref{sec:prelim-prefix} to be
$q_{\mathrm{old}}$ and $r_{\mathrm{exec}}$, two evaluations of the same
checkpoint along different numerical paths.  For a realized history $h_t$ and
current token $x_t$,
\begin{equation}
\label{eq:history-factor}
\frac{P_{q_{\mathrm{old}}}(h_t,x_t)}{P_{r_{\mathrm{exec}}}(h_t,x_t)}
=
\underbrace{\frac{P_{q_{\mathrm{old}}}(h_t)}{P_{r_{\mathrm{exec}}}(h_t)}}_{\text{history factor}}
\cdot
\underbrace{\frac{q_{\mathrm{old}}(x_t\mid h_t)}{r_{\mathrm{exec}}(x_t\mid h_t)}}_{\text{current-token factor}} .
\end{equation}
\Eqref{eq:history-factor} isolates the term omitted by token-local
correction: existing token-local rules retain only the current-token factor
\citep{qiu_2026_fp8rl,zhou_2026_ais}.  A current-token factor near one
does not imply a history factor near one.  Earlier actions determine the
current observation and reachable state.

\begin{figure}[t]
  \centering
  \begin{minipage}[t]{0.46\linewidth}
    \centering
    \includegraphics[width=\linewidth]{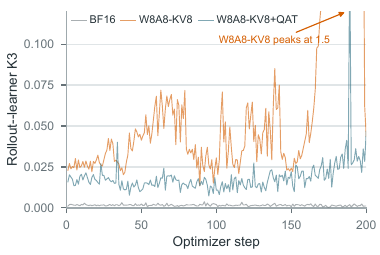}
    \caption{W8A8-KV8+QAT reduces rollout--learner K3 relative to unaligned
    W8A8-KV8 at shared weights, but does not eliminate it. Each curve compares
    the learner and executor at matched checkpoints.}
    \label{fig:qat-gap}
  \end{minipage}\hfill
  \begin{minipage}[t]{0.52\linewidth}
    \centering
    \includegraphics[width=\linewidth]{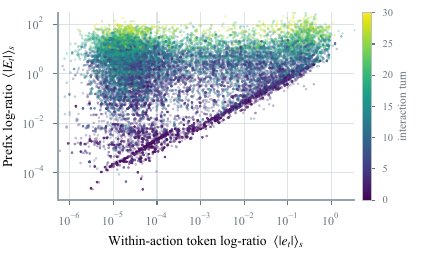}
    \caption{A current action can look locally aligned despite a mismatched
    state-reaching history. 18,683 action segments from ten consecutive updates
    continued from an aligned checkpoint; color encodes interaction turn.}
    \label{fig:prefix-depth}
  \end{minipage}
\end{figure}

Let $m_t=1$ mark policy tokens and $m_t=0$ observations or other non-policy
positions.  After an implementation safety clip, InterQ-RL accumulates token
log-ratios only over policy positions:
\begin{equation}
\label{eq:prefix-weight}
\begin{aligned}
e_t &:= \operatorname{clip}(\delta_t,-B,B), &
n_t &:= \sum_{u\le t}m_u, \\
E_t &:= \sum_{\substack{u\le t\\ m_u=1}}e_u, &
\overline W_t &:= \exp\!\left[\operatorname{clip}\!\left(E_t,-\epsilon_{\mathrm{low}}\sqrt{n_t},\epsilon_{\mathrm{high}}\sqrt{n_t}\right)\right].
\end{aligned}
\end{equation}
The mask excludes environment feedback from the log-ratio sum while retaining
earlier action factors in later state-reaching prefixes, matching the
policy-position scope of the exact product in \S\ref{sec:prelim-prefix}.  To
keep clipping strength comparable across policy positions, we scale the
log-space envelope with $\sqrt{n_t}$
\citep{zhang_2026_ctpo}; here it bounds the correction for residual mismatch
between the two forward paths.  We reuse its envelope parameters,
$\epsilon_{\mathrm{low}}=0.025$ and $\epsilon_{\mathrm{high}}=0.05$; the
per-token log-ratio clip uses verl's default numerical-safety bound $B$.
The weight $\overline W_t$ enters as a detached coefficient, so
gradients flow only through the current-token score function.

Figure~\ref{fig:prefix-depth} shows current-action and state-reaching prefix
ratios separating with interaction depth, motivating correction over the
policy history rather than the current token alone.

\subsection{Allocating Correction by Support, Not Magnitude}
\label{sec:method-alloc}

Let an action segment $s$ be a maximal contiguous run of $m_t=1$ with $n_s$
tokens.  For a segment with $n_s>1$, define
\begin{equation}
\label{eq:segment-stats}
\begin{aligned}
\mu_s &:= \frac1{n_s}\sum_{t\in s}\overline W_t, &
\nu_s &:= \frac1{n_s}\sum_{t\in s}\overline W_t^2, &
\operatorname{ESS}_s &:= \frac{\left(\sum_{t\in s}\overline W_t\right)^2}{\sum_{t\in s}\overline W_t^2}, \\
\rho_s &:= \left[\frac{\operatorname{ESS}_s-1}{n_s-1}\cdot\frac{\nu_s}{\nu_s+\varepsilon_{\mathrm{mom}}}\right]_{[0,1]}, &
C_s &:= \frac1{n_s}\sum_{t\in s}\left(\overline W_t-1\right)^2 . & &
\end{aligned}
\end{equation}
Here $C_s$ measures correction magnitude as mean squared displacement from
identity.  The score $\rho_s$ summarizes prefix-weight concentration within
an action segment through length-normalized ESS; higher values indicate
broader support.  InterQ-RL uses $\rho_s$ as the allocation priority.  The
second-moment factor stabilizes near-zero moments, and uniform weights give
$\rho_s\approx1$.  Segment normalization treats each action as one unit
rather than favoring longer text.  The score follows normalized ESS
\citep{elvira_2022_ess,zhou_2026_ais}.

This distinction occurs in practice: Appendix~B gives a matched pair of
24-token segments with $C_s=0.128882$ and $0.128889$, a difference of only
$7\times10^{-6}$.  Their six largest squared-displacement contributions
carry $76\%$ and $26\%$ of the total, respectively; the latter is close to
the uniform share of $6/24=25\%$.

To set the overall correction strength, InterQ-RL first
defines a uniform reference.  Let $\mathcal S$ contain all action segments in
one rollout/update batch, and define
\begin{equation}
\label{eq:alpha-g}
\mu_g:=\frac1{|\mathcal S|}\sum_{s\in\mathcal S}\mu_s,
\qquad
\nu_g:=\frac1{|\mathcal S|}\sum_{s\in\mathcal S}\nu_s,
\qquad
\alpha_g:=\operatorname{clip}\!\left(\frac{\mu_g}{\sqrt{\nu_g+\varepsilon_{\mathrm{mom}}}},0,1\right).
\end{equation}
Up to the clip and $\varepsilon_{\mathrm{mom}}$,
$\alpha_g^2=\mu_g^2/\nu_g$, the moment ratio underlying relative ESS.
The scalar $\alpha_g$ sets the uniform-reference strength.  InterQ-RL
redistributes this strength by solving

\begin{equation}
\label{eq:alloc}
\max_{\{0\le\alpha_s\le1\}}\ \sum_{s\in\mathcal S}C_s\rho_s\alpha_s
\qquad
\mathrm{subject\ to}
\qquad
\sum_{s\in\mathcal S}C_s\alpha_s^2\le\alpha_g^2\sum_{s\in\mathcal S}C_s .
\end{equation}
The utility $C_s\rho_s\alpha_s$ favors retaining correction with broader
support, while $C_s\alpha_s^2$ measures its resulting displacement.  The
constraint fixes total segment-normalized displacement to the uniform
reference; at the optimum it saturates, so allocation redistributes rather
than increases correction strength (Appendix~B).

\paragraph{Proposition 1.}
If every segment with $n_s>1$ satisfies $C_s>0$ and $\rho_s\in(0,1]$, and
$\alpha_g\in(0,1)$, then \eqref{eq:alloc} has the unique solution
$\alpha_s=\min(c\rho_s,1)$, where $c>0$ is determined uniquely by the budget
equality and can be found by one-dimensional bisection.  If all $\rho_s$ are
equal then $\alpha_s=\alpha_g$ and InterQ-RL reduces exactly to the uniform
reference.  The proof and the $n_s=1$ case are given in Appendix~B.

Higher-$\rho_s$ segments therefore retain more prefix correction, whereas
lower-$\rho_s$ segments are pulled toward identity.  Since
$0\le\alpha_s\le1$, every final weight lies between 1 and $\overline W_t$.

\subsection{The InterQ-RL Objective}
\label{sec:method-objective}

For a fixed executor batch, rewards and group-relative advantages are
detached.
Let $\operatorname{seg}(t)$ denote the action segment containing token $t$,
and let $\ell_t^{\mathrm{base}}(\theta)$ be the
token-level loss of the underlying PPO/GRPO objective.  InterQ-RL defines
the empirical objective
\begin{equation}
\label{eq:objective}
W_t := 1+\alpha_{\operatorname{seg}(t)}\left(\overline W_t-1\right),
\qquad
\mathcal L_{\mathrm{InterQ\text{-}RL}}
:=
\operatorname{Agg}_{\mathrm{base}}\left[m_t\operatorname{stopgrad}(W_t)
\ell_t^{\mathrm{base}}(\theta)\right].
\end{equation}
Equation~\ref{eq:prefix-identity} gives the exact
$q_{\mathrm{old}}/r_{\mathrm{exec}}$ ratio for the state-reaching policy
prefix associated with each token; $W_t$ is its bounded, support-allocated
counterpart.  Equation~\ref{eq:objective} reweights only the base token loss:
$W_t$ accounts for residual execution mismatch after alignment, while the
current-policy PPO/GRPO ratio, clipping rule, executor-sampled advantage, and
aggregation remain unchanged.  With no mismatch, $W_t=1$ and the objective
reduces to the base PPO/GRPO objective.  Prefix sums, segment statistics, and
the one-dimensional solver are linear in policy-token count and reuse
existing rollout and learner log probabilities, requiring no additional
model forward pass.  Algorithm~\ref{alg:interq} gives the per-batch procedure.

\begin{algorithm}[H]
\caption{Execution-prefix correction with support-aware allocation.  The
returned $W_t$ is the coefficient applied in \eqref{eq:objective}.  Constants:
$\epsilon_{\mathrm{low}}=0.025$, $\epsilon_{\mathrm{high}}=0.05$,
$\varepsilon_{\mathrm{mom}}=10^{-8}$, and 80 bisection steps.}
\label{alg:interq}
\begin{algorithmic}[1]
\STATE $e_t\gets\operatorname{clip}\!\left(\log q_{\mathrm{old}}(x_t\mid h_t)-\log r_{\mathrm{exec}}(x_t\mid h_t),\,-B,\,B\right)$
\STATE $E_t\gets\sum_{u\le t}m_ue_u$ \quad(state-reaching policy prefix)
\STATE $n_t\gets\sum_{u\le t}m_u$
\STATE $\overline W_t\gets\exp\!\left[\operatorname{clip}\!\left(E_t,-\epsilon_{\mathrm{low}}\sqrt{n_t},\epsilon_{\mathrm{high}}\sqrt{n_t}\right)\right]$
\STATE $\mathcal S\gets$ maximal runs of consecutive $m_t=1$ \quad(the action segments)
\STATE For each $s\in\mathcal S$, compute $\mu_s,\nu_s,\rho_s,C_s$ from \eqref{eq:segment-stats}
\STATE $\alpha_g\gets\operatorname{clip}\!\left(\mu_g/\sqrt{\nu_g+\varepsilon_{\mathrm{mom}}},0,1\right)$ \quad(batch means $\mu_g,\nu_g$ from \eqref{eq:alpha-g})
\STATE $c\gets$ root of $\sum_sC_s\min(c\rho_s,1)^2=\alpha_g^2\sum_sC_s$, by bisection
\STATE $\alpha_s\gets\min(c\rho_s,1)$ \quad(Appendix~\ref{app:premises} for $n_s=1$)
\STATE $W_t\gets1+\alpha_{\operatorname{seg}(t)}\left(\overline W_t-1\right)$
\STATE \textbf{return} $W_t$
\end{algorithmic}
\end{algorithm}

\section{Experiments}
\label{sec:experiments}

\subsection{Experimental Setup}
\label{sec:exp-setup}

\paragraph{Settings and comparisons.}
We evaluate on three multi-turn agent environments: ALFWorld
\citep{shridhar_2021_alfworld}, WebShop \citep{yao_2022_webshop}, and
ScienceWorld \citep{wang_2022_scienceworld}.
ALFWorld and WebShop both use Qwen3-4B-Instruct-2507 and report on 134
\texttt{valid\_unseen} tasks and 500 official test tasks, respectively.
Because Qwen3-4B provides limited learning signal on the more demanding
ScienceWorld setting, we use Qwen3-8B and Qwen3-30B-A3B-Instruct-2507,
covering dense and MoE backbones
\citep{yang_2025_qwen3}; endpoint evaluation uses 1,819 official test
instances.  All runs use GRPO \citep{shao_2024_deepseekmath}.
Table~\ref{tab:endpoint} is the primary end-to-end comparison: the published
baselines retain their objectives; all arms share task IDs, evaluator,
training horizon, and optimizer settings; and all quantized arms share the
W8A8-KV8 rollout.  Baselines are BF16, Flash-RL with token-local TIS
\citep{yao_2025_rollout_mismatch}, AIS \citep{zhou_2026_ais}, QuRL with ACR+UAQ
\citep{li_2026_qurl}, and QAT with W/A+KV alignment.
Flash-RL uses verl's built-in rollout importance-sampling path; AIS and
QuRL are reimplemented from their papers.  Appendix~C gives method-specific
settings.

\paragraph{Protocol and metrics.}
All settings train for 400 optimizer steps. ALFWorld and WebShop endpoints are
the mean of the final six checkpoint evaluations.  ScienceWorld reports the
full official test at a fixed endpoint checkpoint.  We
report the ALFWorld success rate, the WebShop score and success rate, and
the ScienceWorld macro normalized score and per-episode solved rate.  Success
and solved rates count fully completed tasks, whereas the WebShop score and
the ScienceWorld macro score credit partial progress.
To separate endpoint quality from learning speed, we report normalized AULC:
the trapezoidal area under the evaluation curve, with scores recorded every
10 steps and the area divided by the common 400-step horizon.  ALFWorld uses
the same \texttt{valid\_unseen} set, WebShop the same 500-instruction
official test set, and ScienceWorld the 259-instance development monitor.
Interaction cost is the mean number of turns over successful endpoint
episodes; cap exhaustion counts unsuccessful episodes that consume the
environment's full action budget.
Appendix~C specifies baseline rules, data splits, evaluation protocols,
hardware, and quantization.

\subsection{End-to-End Results}
\label{sec:exp-main}

\begin{table}[!t]
\caption{End-to-end comparison of complete methods under a common
  training budget.
  ALFWorld and WebShop report the mean over the final six
  checkpoint evaluations; ScienceWorld reports the full official test at a
  fixed endpoint checkpoint.  ALFWorld and WebShop use Qwen3-4B; ScienceWorld
  headers give the backbone size.  Bold marks the best quantized result.
  $\dagger$ QuRL collapsed at step 278 and has no common-horizon endpoint.}
\label{tab:endpoint}
\centering
\footnotesize
\setlength{\tabcolsep}{2.0pt}
\begin{tabular*}{\linewidth}{@{\extracolsep{\fill}}lccccccc@{}}
\toprule
& ALFWorld & \multicolumn{2}{c}{WebShop} &
\multicolumn{2}{c}{ScienceWorld (8B)} &
\multicolumn{2}{c}{ScienceWorld (A3B)} \\
\cmidrule(lr){2-2}\cmidrule(lr){3-4}\cmidrule(lr){5-6}\cmidrule(l){7-8}
Method & Success $\uparrow$ & Score $\uparrow$ & Success $\uparrow$ &
Macro $\uparrow$ & Solved $\uparrow$ & Macro $\uparrow$ & Solved $\uparrow$ \\
\midrule
BF16 & 97.1 & 78.0 & 54.0 & 66.9 & 74.1 & 67.0 & 74.3 \\
\midrule
Flash-RL & 64.1 & 72.8 & 44.4 & 63.2 & 69.2 & 65.8 & 71.5 \\
AIS & 92.3 & 76.5 & 51.5 & 59.6 & 64.2 & 65.0 & 70.6 \\
QuRL & Coll.$^\dagger$ & 75.3 & 51.7 & 57.1 & 63.9 & 64.0 & 69.6 \\
QAT & 90.4 & 74.9 & 49.7 & 65.5 & 75.8 & 69.3 & 76.6 \\
InterQ-RL & \textbf{96.4} & \textbf{77.7} & \textbf{53.5} &
\textbf{67.8} & \textbf{76.1} & \textbf{69.9} & \textbf{76.9} \\
\bottomrule
\end{tabular*}
\end{table}

\begin{figure}[!t]
  \centering
  \includegraphics[width=\linewidth]{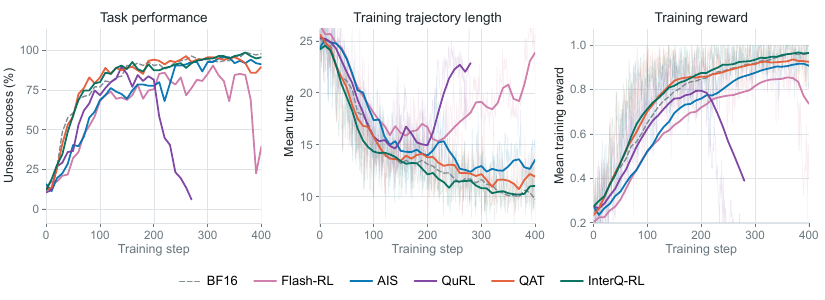}
  \caption{On ALFWorld, InterQ-RL remains stable while QAT rolls back,
  Flash-RL destabilizes, QuRL collapses, and AIS learns more slowly. Mean turns
  in the middle panel are averaged over all training trajectories. For
  trajectory length and reward, smoothed trends use 10-step block medians
  followed by light EMA.}
  \label{fig:dynamics}
\end{figure}

\paragraph{Endpoint performance and training dynamics.}
InterQ-RL gives the strongest quantized result on every metric in
Table~\ref{tab:endpoint}, including the four that count fully completed
tasks.  It improves over QAT by 6.0 and 2.8 points on ALFWorld success
and WebShop score and by 2.3 and 0.6 on the two ScienceWorld macro scores,
and over the strongest of Flash-RL, AIS, and QuRL by 1.2 to 6.9 points.
It matches BF16 within 0.7 points on ALFWorld success, 0.3 on WebShop score,
and 0.5 on WebShop success, and exceeds it by 0.9 and 2.9 on the ScienceWorld
macro
scores and 2.0 and 2.6 on the solved rates.
Figure~\ref{fig:dynamics} places the ALFWorld endpoint in its training
context.  QAT reaches 95.5\% at step 300 but falls to 89.6\% by step
400, while InterQ-RL is still at 95.5\%.  The endpoint separation shows that
interaction correction sustains task performance.  Flash-RL drops from 84.3\% at
step 370 to 39.6\% at step 400.  QuRL falls to 6.0\% at step 270 as its
training reward falls from a peak of 0.992 to 0.141, and terminates at step
278.  AIS avoids collapse but learns more slowly and unevenly: it reaches
69.4\% at step 150 versus 88.1\% for InterQ-RL, drops from 79.9\% at step 210
to 67.9\% ten steps later, and reaches a 92.3\% last-six endpoint versus
96.4\% for InterQ-RL.

\paragraph{Learning speed and interaction cost.}

\begin{table}[!t]
\caption{Learning speed and the interaction cost of successful
  trajectories.  ALFWorld and WebShop use Qwen3-4B; ScienceWorld headers give
  the backbone size.  Turns average over successful episodes only; bold marks
  the best quantized AULC.  QuRL collapsed on ALFWorld at step 278 and has no
  result on the common horizon.}
\label{tab:aulc}
\centering
\footnotesize
\setlength{\tabcolsep}{2.0pt}
\begin{tabular*}{\linewidth}{@{\extracolsep{\fill}}lcccccccc@{}}
\toprule
& \multicolumn{2}{c}{ALFWorld} & \multicolumn{2}{c}{WebShop} &
\multicolumn{2}{c}{ScienceWorld (8B)} &
\multicolumn{2}{c}{ScienceWorld (A3B)} \\
\cmidrule(lr){2-3}\cmidrule(lr){4-5}\cmidrule(lr){6-7}\cmidrule(l){8-9}
Method & AULC $\uparrow$ & Turns $\downarrow$ &
AULC $\uparrow$ & Turns $\downarrow$ &
AULC $\uparrow$ & Turns $\downarrow$ &
AULC $\uparrow$ & Turns $\downarrow$ \\
\midrule
BF16 & 82.1 & 10.98 & 75.2 & 5.23 & 56.4 & 20.45 & 54.1 & 19.04 \\
\midrule
Flash-RL & 64.9 & 20.34 & 72.5 & 5.09 & 47.6 & 31.93 & 52.1 & 20.31 \\
AIS & 72.5 & 12.43 & 72.8 & 6.61 & 45.4 & 22.32 & 49.6 & 18.07 \\
QuRL & --- & --- & 72.2 & 6.23 & 50.8 & 21.19 & 45.5 & 32.32 \\
QAT & \textbf{82.3} & 12.72 & 72.3 & 6.51 & 54.4 & 22.38 & 54.1 & 20.20 \\
InterQ-RL & 82.0 & 10.06 & \textbf{73.3} & 4.60 &
\textbf{54.6} & 21.20 & \textbf{57.4} & 18.18 \\
\bottomrule
\end{tabular*}
\end{table}

In Table~\ref{tab:aulc}, InterQ-RL has the highest quantized AULC in three
settings and is within 0.33 of QAT on ALFWorld
(81.96 versus 82.29), while remaining above Flash-RL, AIS, and QuRL
in all four.  While these three baselines are 2.0--17.2 points below BF16
across common-horizon results, InterQ-RL stays within 1.9 points in the first
three settings and exceeds BF16 by 3.3 on ScienceWorld/30B-A3B.
InterQ-RL also reduces interaction cost.  It yields the
shortest successful trajectories on ALFWorld and WebShop, reducing the
QAT averages from 12.72 to 10.06 and from 6.51 to 4.60 turns,
respectively.  On the 117 ALFWorld tasks solved by both methods, InterQ-RL
averages 9.64 turns versus 12.79 for QAT.  In both ScienceWorld
settings, it remains within 0.11 turns of the shortest quantized arm.  Cap
exhaustion also falls from 10.45\% to 4.48\% on ALFWorld and from 1.60\% to
0.40\% on WebShop.  On ALFWorld and both ScienceWorld settings, the arm with
the lowest AULC also has the longest successful trajectories, whereas
InterQ-RL combines higher AULC with shorter trajectories.

\begin{figure}[!t]
  \centering
  \includegraphics[width=\linewidth]{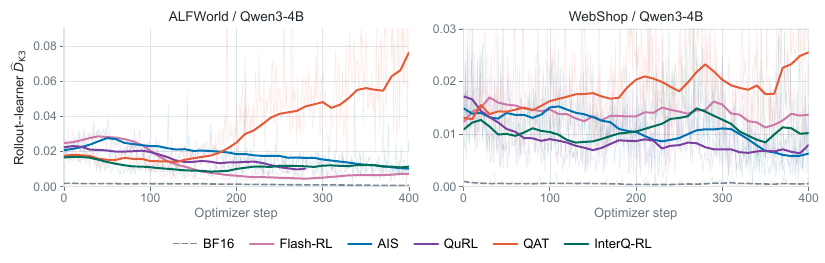}
  \caption{During continued optimization, K3 grows under QAT and
  remains stable under InterQ-RL. K3 is computed on the
  trajectories of each arm and shown as smoothed trends. The QuRL curve on
  ALFWorld ends at step 278.}
  \label{fig:k3}
\end{figure}

\paragraph{Rollout--learner K3.}
Over the first 100 ALFWorld steps, QAT and InterQ-RL have similar median
K3 (0.0155 versus 0.0125); by steps 301--400 they separate to 0.0598 and
0.0119 (Figure~\ref{fig:k3}).  WebShop shows the same pattern, with
QAT and InterQ-RL ending at 0.0210 and 0.0081.  On each arm's rollouts,
QAT's rollout--learner gap widens, whereas InterQ-RL's remains
stable.
A low K3 alone does not imply strong task performance: QuRL's K3 falls as it
collapses, while Flash-RL and AIS end with K3 values of 0.0071 and 0.0107
despite lower success (64.1\% and 92.3\%) than InterQ-RL (0.0119 at 96.4\%);
AIS also has lower AULC (72.5 versus 82.0).

\paragraph{Additional design analyses.}
Appendix~\ref{app:design-analyses} provides two additional analyses on
ALFWorld.
Prefix applies the bounded prefix weights directly, Global uses one
coefficient across all segments, and InterQ-RL allocates segment-specific
coefficients by effective support (Figure~\ref{fig:ablation}).  Prefix and
Global reach 88.8\% and 94.8\% success at steps 290 and 230 but end at 37.3\%
and 22.4\% at step 400, whereas InterQ-RL ends at 95.5\%.  Global and
InterQ-RL use the same segment-normalized displacement budget.  These results
show that prefix correction provides useful learning signal, while
support-aware allocation sustains late-stage performance under a fixed
correction budget.
As an additional ablation, Table~\ref{tab:correction} pairs Flash-RL's TIS
rule and the AIS correction rule with the same QAT learner used by InterQ-RL.
QAT+TIS and QAT+AIS reach 90.55\% and 92.66\% last-six success, compared
with 90.42\% for QAT alone, while their AULCs fall from 82.29 to 70.59 and
76.27.  Combining QAT with token-local correction therefore does not resolve
the residual mismatch in multi-turn agent RL.  InterQ-RL reaches 96.39\%
success with 81.96 AULC and lowers cap exhaustion to 4.48\%, showing the
effectiveness of our method for quantized agent RL.

\section{Related Work}
\label{sec:related}

\paragraph{Quantized LLM reinforcement learning.}
Quantized LLM RL builds on weight, activation, and KV-cache quantization
\citep{dettmers_2022_llmint8,xiao_2023_smoothquant,liu_2024_kivi,deepseekai_2025_deepseekv3},
and spans learner--rollout
alignment and mismatch correction.  FP8-RL quantizes rollout W/A and the KV
cache \citep{qiu_2026_fp8rl}.  QaRL and Jet-RL align the learner forward
with quantized rollout \citep{gu_2026_qarl,xi_2026_jetrl}.  Flash-RL, AIS, and QuRL
control training--inference mismatch through clipping, adaptive weighting, or
update-aware quantization
\citep{yao_2025_rollout_mismatch,zhou_2026_ais,li_2026_qurl},
all in single-turn generation.
InterQ-RL addresses residual execution mismatch along the policy histories of
multi-turn agents.

\paragraph{Policy optimization under distribution shift.}
Off-policy RL corrects distribution shift through importance weighting
\citep{precup_2000_offpolicy,munos_2016_retrace,espeholt_2018_impala}.  LLM
methods span response-level objectives, cumulative-token ratios, prefix
surrogates, and whole-sequence weighting or masking
\citep{zheng_2025_gspo,zhang_2026_ctpo,lei_2026_minpro,zhang_2026_pnpo,li2026trustregionmaskinglonghorizon}.
These methods address general rollout or policy-version mismatch at different
granularities.  InterQ-RL instead targets quantization-induced mismatch that
remains after alignment and allocates bounded correction across action segments
by effective support under a fixed displacement budget.

\paragraph{Agentic reinforcement learning.}
GiGPO and LOOP extend policy optimization to multi-turn interaction,
targeting credit assignment and training efficiency
\citep{feng_2025_gigpo,chen_2025_longhorizon}; Agent-R1, AgentGym-RL, and
SkyRL-Agent provide open training frameworks
\citep{cheng_2026_agentr1,xi_2026_agentgymrl,cao_2025_skyrl_agent}.
InterQ-RL addresses a complementary execution-level source of distribution
shift: mismatch introduced when the behavior policy runs in low precision and
propagated through action-conditioned history.

\section{Conclusion}

We identified interaction-structured execution mismatch in quantized agentic
RL: even after W/A+KV alignment, residual rollout--learner mismatch compounds
along the state-reaching policy history.  InterQ-RL
applies bounded execution-prefix correction along this history and allocates
correction strength across action segments by effective support under a fixed
displacement budget, reusing
existing log probabilities without another forward pass.  Across four
settings spanning three environments and 4B--30B dense and MoE backbones
under a common W8A8-KV8 rollout, InterQ-RL achieves the strongest
quantized endpoint result in every setting, matches BF16 to within 0.7
points and exceeds it in two.  Its AULC is within 0.33 of QAT on
ALFWorld and highest among quantized arms in the other three settings.


% ...
%
% ...

\bibliography{iclr2027_conference}
\bibliographystyle{iclr2027_conference}

\appendix

\clearpage

\section{Quantization Alignment and Execution Diagnostics}
\label{app:align}

\subsection{Linear weight and activation path}
\label{app:linear}

The rollout executor quantizes linear operators to FP8 E4M3 with block
scaling: weight scales cover $128\times128$ blocks and activation scales cover
$1\times128$ blocks.  The learner forward applies the matching
quantize--dequantize rule to the same operands, while master parameters,
gradients, and optimizer states stay in high precision.  Scales are stored in
FP32.  Quantization covers every linear layer except \texttt{lm\_head}, the
embedding, and the MoE gate.

\subsection{KV cache path}
\label{app:kv}

The rollout executor stores the KV cache in \texttt{fp8\_per\_token\_head}
format.  For the K or V vector $x$ at a given token and attention head, the
cache-write kernel computes
\begin{equation}
\label{eq:kv-quant}
s(x):=\max\!\left(\frac{\max_j|x_j|}{448},\,10^{-6}\right),
\qquad
\widehat x:=s(x)\,\operatorname{FP8}_{\mathrm{E4M3}}\!
\left[\operatorname{clip}\!\left(\frac{x}{s(x)},-448,448\right)\right].
\end{equation}
The scale depends on the current token--head slice alone.  The learner applies the
same quantize--dequantize rule at the corresponding cache-write sites: keys
after key normalization and RoPE, values at the matching V cache-write
position.  The learner rule matches the vLLM cache-write kernel bitwise, and
its straight-through backward is the identity.

\subsection{The K3 diagnostic}
\label{app:k3}

For an executor sample $x_t\sim r_{\mathrm{exec}}(\cdot\mid h_t)$, the
token log-ratio used to quantify the residual mismatch is
$\delta_t:=\log q_{\mathrm{old}}(x_t\mid h_t)-\log r_{\mathrm{exec}}(x_t\mid h_t)$.
We report the standard K3 estimator
\citep{schulman_2020_klapprox,shao_2024_deepseekmath},
\begin{equation}
\label{eq:k3}
\widehat D_{\mathrm{K3}}
:=
\frac{1}{\sum_t m_t}\sum_{t:\,m_t=1}\left[\exp(\delta_t)-\delta_t-1\right],
\end{equation}
in which every term is nonnegative because $\exp(\delta_t)\ge1+\delta_t$.  On
samples from $r_{\mathrm{exec}}$,
\begin{equation}
\label{eq:k3-expectation}
\mathbb E_{r_{\mathrm{exec}}}\!\left[\exp(\delta_t)\right]=1,
\qquad
-\,\mathbb E_{r_{\mathrm{exec}}}\!\left[\delta_t\right]
=
D_{\mathrm{KL}}\!\left(r_{\mathrm{exec}}(\cdot\mid h_t)\,\Vert\,q_{\mathrm{old}}(\cdot\mid h_t)\right),
\end{equation}
so the expectation of \eqref{eq:k3} is
$D_{\mathrm{KL}}(r_{\mathrm{exec}}\Vert q_{\mathrm{old}})$, the
rollout--learner divergence at a shared checkpoint.

\clearpage

\section{Support-Aware Allocation}
\label{app:alloc}

\subsection{The displacement budget}
\label{app:budget}

Fix one rollout/update batch and let $\mathcal S$ be its action segments, with
$\mu_s$, $\nu_s$, $\operatorname{ESS}_s$, $\rho_s$, and $C_s$ as in
\eqref{eq:segment-stats}, which applies to segments with $n_s>1$ and takes
$\varepsilon_{\mathrm{mom}}=10^{-8}$.  With
$W_t=1+\alpha_s(\overline W_t-1)$ from \eqref{eq:objective}, each segment
satisfies
\begin{equation}
\label{eq:segment-displacement}
\frac1{n_s}\sum_{t\in s}(W_t-1)^2
=
\alpha_s^2\,\frac1{n_s}\sum_{t\in s}\left(\overline W_t-1\right)^2
=
C_s\alpha_s^2 .
\end{equation}
The constrained quantity in \eqref{eq:alloc} is therefore
$\sum_{s}C_s\alpha_s^2=\sum_{s}n_s^{-1}\sum_{t\in s}(W_t-1)^2$, the
segment-normalized squared displacement of the applied weights from the
identity weight 1.  Giving every segment $\alpha_s=\alpha_g$ makes this
quantity $\alpha_g^2\sum_sC_s$, which is exactly the budget.  Support-aware
allocation and the uniform reference therefore work under the same
segment-normalized displacement budget.

\paragraph{Weight concentration and correction magnitude.}
For the positive weights $\overline W_t$ within a segment,
\begin{equation}
\frac{\operatorname{ESS}_s}{n_s}
=
\frac{\mu_s^2}{\nu_s}
=
\frac{1}{1+\operatorname{CV}_s^2},
\qquad
\operatorname{CV}_s^2:=\frac{\nu_s-\mu_s^2}{\mu_s^2}.
\end{equation}
Weight concentration therefore lowers normalized ESS and increases
$\operatorname{CV}_s^2$.  The finite-length factor in $\rho_s$ preserves this
concentration ordering, while
$\nu_s/(\nu_s+\varepsilon_{\mathrm{mom}})$ stabilizes a near-zero moment.
The separate quantity $C_s$ records correction magnitude away from identity.

\subsection{Equal magnitude, different support}
\label{app:pair}

\begin{figure}[H]
  \centering
  \includegraphics[width=\linewidth]{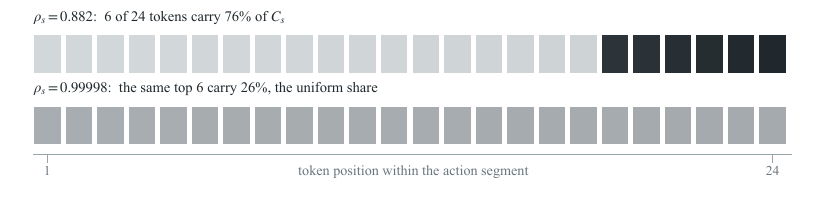}
  \caption{Two action segments with indistinguishable $C_s$ and different
  $\rho_s$.  Each block is one token, shaded by that
  token's share of squared displacement within its segment.  The two
  values of $C_s$ differ by $7\times10^{-6}$.}
  \label{fig:support-pair}
\end{figure}

Using the replay underlying Figure~\ref{fig:prefix-depth},
Figure~\ref{fig:support-pair} shows how the two 24-token segments of
\S\ref{sec:method-alloc} distribute their correction across tokens.

$C_s$ therefore does not determine $\rho_s$.  We selected the pair by
searching for a near-tie in $C_s$.

\subsection{Closed-form allocation}
\label{app:proof}

\paragraph{Proposition 1 (restated from \S\ref{sec:method-alloc}).}
Let every segment with $n_s>1$ satisfy $C_s>0$ and $\rho_s\in(0,1]$, and let
$\alpha_g\in(0,1)$.  Consider
\begin{equation}
\label{eq:alloc-restated}
\max_{\alpha\in[0,1]^{|\mathcal S|}}\ \sum_{s\in\mathcal S}C_s\rho_s\alpha_s
\qquad
\mathrm{subject\ to}
\qquad
\sum_{s\in\mathcal S}C_s\alpha_s^2\le\alpha_g^2\sum_{s\in\mathcal S}C_s .
\end{equation}
Then the optimum exists, is unique, and has the form
\begin{equation}
\label{eq:alloc-solution}
\alpha_s=\min(c\rho_s,1),
\qquad\text{where $c>0$ is the unique root of }
\sum_{s}C_s\min(c\rho_s,1)^2=\alpha_g^2\sum_{s}C_s .
\end{equation}
If all $\rho_s$ are equal, then $\alpha_s=\alpha_g$.

\paragraph{Proof.}
Substitute
$\beta_s:=\sqrt{C_s}\,\alpha_s$, $v_s:=\sqrt{C_s}\,\rho_s$,
$u_s:=\sqrt{C_s}$, and $R^2:=\alpha_g^2\sum_sC_s$, which turns
\eqref{eq:alloc-restated} into
\begin{equation}
\label{eq:alloc-ball}
\max_\beta\ \langle v,\beta\rangle
\qquad
\mathrm{subject\ to}
\qquad
\|\beta\|_2\le R,
\quad
0\le\beta_s\le u_s .
\end{equation}
The feasible set is the intersection of a Euclidean ball with a box, hence
convex and compact, so an optimum exists.  Because $v_s>0$ and
$R^2<\sum_su_s^2$, any feasible point with $\|\beta\|_2<R$ has a coordinate
strictly below its box bound, and increasing $\beta_s$ along that coordinate
strictly increases the objective.  Every optimum therefore satisfies
$\|\beta\|_2=R$ and exhausts the budget.

Write the KKT conditions for \eqref{eq:alloc-ball}.  The ball constraint is
active, so its multiplier is positive, and the coordinatewise optimality
condition gives $\beta_s=\min(c\,v_s,u_s)$ for a constant $c>0$ fixed by that
multiplier.  In the original variables this condition is
\eqref{eq:alloc-solution}.  Now set $g(c):=\sum_sC_s\min(c\rho_s,1)^2$.
Since every $\rho_s>0$, the function $g$
is continuous and strictly increasing from $g(0)=0$ until it saturates at
$\sum_sC_s$.  Because $0<R^2=\alpha_g^2\sum_sC_s<\sum_sC_s$, exactly one $c$
satisfies $g(c)=R^2$, which also establishes uniqueness of the optimum.  If
all $\rho_s$ equal a common $\rho>0$, then all $\alpha_s$ coincide, and the
budget equality gives $\alpha_s=\alpha_g$.\hfill$\square$

\subsection{Allocation over a full training run}
\label{app:training}

\begin{figure}[H]
  \centering
  \includegraphics[width=\linewidth]{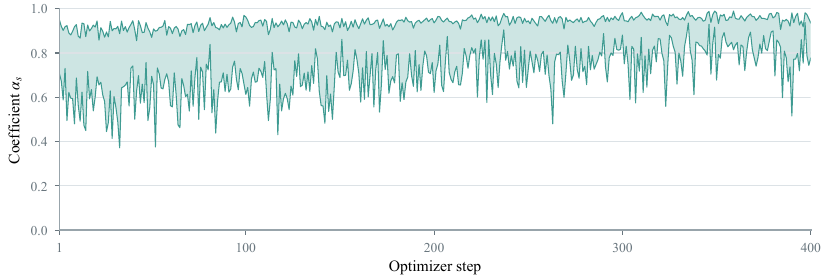}
  \caption{Range of the allocated $\alpha_s$ within each batch over one
  complete 400-step ALFWorld run: the band spans the smallest to the largest
  $\alpha_s$ in that batch.  Across the 400 batches the median range is 0.717
  to 0.942 with median width 0.215; over the whole run $\alpha_s$ reaches
  0.372 and 0.989.}
  \label{fig:allocation-range}
\end{figure}

The range of $\alpha_s$ within a batch never collapses: over one complete
400-step ALFWorld run, its width stays above 0.01
(Figure~\ref{fig:allocation-range}).  The range shifts upward and narrows over
training, from a median range of 0.634--0.915 with width 0.287 over steps
1--100 to 0.810--0.959 with width 0.151 over steps 301--400.

\subsection{Premises of Proposition~1}
\label{app:premises}

Since \eqref{eq:prefix-weight} defines $\overline W_t$ as an exponential,
$\overline W_t>0$.  Positivity makes $\operatorname{ESS}_s>1$ strict for
$n_s>1$, so $\rho_s>0$, and makes $\mu_g>0$, so $\alpha_g>0$.  Jensen's
inequality gives $\nu_s\ge\mu_s^2$ for each segment, and a second application
after averaging over the batch gives $\nu_g\ge\mu_g^2$, hence $\alpha_g<1$.

Every one of the 18,683 segments in the replay underlying
Figure~\ref{fig:prefix-depth} has $C_s>0$; $C_s=0$ would require
$\overline W_t=1$ at all of a segment's tokens, and the smallest segment-mean
$|\delta_t|$ there is $5.3\times10^{-7}$.

For $n_s=1$ the factor $(\operatorname{ESS}_s-1)/(n_s-1)$ is $0/0$ and
\eqref{eq:segment-stats} leaves $\rho_s$ undefined.  We substitute the mean
$\rho_s$ over the segments with $n_s>1$ in the same batch.  Two of the 18,683
segments in that replay are of this kind.

\clearpage

\section{Experimental Setup}
\label{app:setup}

\subsection{Environments, data, and evaluation}
\label{app:envs}

Table~\ref{tab:setup} lists the four environment--backbone settings, their
training horizons, AULC sets, and endpoint sets.

\begin{table}[H]
\centering
\caption{Environments, backbones, training steps, and evaluation sets.}
\label{tab:setup}
\footnotesize
\begin{tabular}{llrll}
\toprule
Environment & Backbone & Steps & AULC set & Endpoint \\
\midrule
ALFWorld & Qwen3-4B & 400 & \texttt{valid\_unseen} (10-step) & \texttt{valid\_unseen} (final-six) \\
WebShop & Qwen3-4B & 400 & official test (10-step) & official test (final-six) \\
ScienceWorld & Qwen3-8B & 400 & 259 monitor (10-step) & official test (step 400) \\
 & Qwen3-30B-A3B & 400 & 259 monitor (10-step) & official test (step 400) \\
\bottomrule
\end{tabular}
\end{table}

All three environments use their public splits.  ALFWorld trains on 3,553
games and uses the 134 \texttt{valid\_unseen} games for its displayed curve,
AULC, and endpoints \citep{shridhar_2021_alfworld}.  WebShop trains on 11,587
instructions and uses the 500 official test instructions for its displayed
curve, AULC, and endpoints
\citep{yao_2022_webshop}.  ScienceWorld trains on the official training
variations of each task and evaluates on the 1,819 official test variations
\citep{wang_2022_scienceworld}.

Scores on the AULC sets are recorded every 10 steps.  ALFWorld and WebShop
endpoints are the mean of the final six checkpoint evaluations on the
endpoint sets in Table~\ref{tab:setup}.  For periodic ScienceWorld
evaluation, we use a fixed set of 259 held-out variations: the first ten by
variation ID from each of the 30 task families, and all of them for the
families that have fewer than ten.  The full official test set of 1,819
\texttt{(task\_name, variation\_id)} pairs is run once, at the final training
step.  Both sets weight task families equally when forming the macro
normalized score.
For both ScienceWorld settings, each action is capped at 32 generated tokens
(\texttt{max\_generation\_tokens}=32).  The Qwen3-8B setting disables
thinking (\texttt{enable\_thinking=false}).

Within an environment, all arms share the training and evaluation task IDs,
the initial checkpoint, the prompt order, the turn cap, and the evaluator.
Training rollouts sample at temperature 1.0 with top-$p$ 1.0 and top-$k$ disabled; all
evaluation decodes greedily.  Each environment supplies its own reward: a
binary success indicator for ALFWorld; for WebShop, the environment's final
reward, with 0.05 subtracted per malformed tool call during training; and for
ScienceWorld, the environment score clipped at zero and divided by 100, where
an episode counts as solved at the full score of 100.

All turns of an episode are trained as one concatenated sequence: the prompt,
each action, and each observation follow one another, and the
policy-gradient loss acts only on action tokens (the mask $m_t$ of
\S\ref{sec:prelim-traj}).  The response budget in Table~\ref{tab:budget}
therefore covers all actions and observations of an episode together.

\begin{table}[H]
\centering
\caption{Sequence budgets.  All entries are token counts except the turn
cap.  Context is the rollout engine's window, Prompt caps the first-turn
prompt, and Response caps the actions and observations appended after it;
when appending the next observation would exceed this budget, the episode
stops at the response boundary.}
\label{tab:budget}
\footnotesize
\begin{tabular}{lrrrr}
\toprule
Environment & Turn cap & Context & Prompt & Response \\
\midrule
ALFWorld & 30 & 24,576 & 2,048 & 20,480 \\
WebShop & 15 & 24,576 & 4,096 & 20,480 \\
ScienceWorld / 8B & 100 & 40,960 & 1,024 & 39,936 \\
ScienceWorld / 30B-A3B & 100 & 50,176 & 1,024 & 49,152 \\
\bottomrule
\end{tabular}
\end{table}

An episode that neither succeeds nor terminates before the turn cap is
reached counts as cap exhaustion.  Reported cap-exhaust values and
successful-episode turns are computed from the same step-400 episode records
as the corresponding single-checkpoint success values.  The training horizon
and sequence budgets in Tables~\ref{tab:setup} and~\ref{tab:budget} are fixed
by the environment and model size and shared across arms within each setting.

\subsection{Training hyperparameters}
\label{app:hparams}

Table~\ref{tab:hparams} gives the common GRPO hyperparameters;
\S\ref{app:configs} describes method-specific objective deviations.

\begin{table}[H]
\centering
\caption{GRPO hyperparameters.}
\label{tab:hparams}
\footnotesize
\begin{tabular}{ll}
\toprule
Hyperparameter & Value \\
\midrule
Advantage estimator & GRPO \\
Optimizer & AdamW \\
Adam settings & $\beta=(0.9,0.999)$; weight decay $0.01$; clip norm $1.0$ \\
LR / schedule & $1\times10^{-6}$; constant; no warmup \\
PPO clip & $\epsilon_{\mathrm{PPO}}=0.2$; dual-clip $c=10$ (except QuRL) \\
KL & penalty in the loss, coefficient 0.001 \\
Loss aggregation & \texttt{seq-mean-token-mean} \\
Responses per prompt & 8 \\
Train batch size & 16 \\
\bottomrule
\end{tabular}
\end{table}

\subsection{Quantization and correction configurations}
\label{app:configs}

Every quantized arm uses the same rollout stack: blockwise FP8 W8A8 linear
operators and a per-token-per-head FP8 KV cache.  QAT and InterQ-RL
reproduce both rules in the learner forward, as specified in
\S\ref{app:align}.  For the ALFWorld design analysis, we additionally combine
Flash-RL's TIS and AIS with this learner configuration while retaining their
correction rules and reported parameters.
InterQ-RL retains verl's default per-token log-ratio safety bound $B=20$.

Baseline implementations follow each method's published objective, using
reported parameter values where specified.  Flash-RL applies token-local TIS
to the rollout--learner ratio with bound $C=2$
\citep{yao_2025_rollout_mismatch}.  AIS keeps a truncated weight, at bound $C=5$,
and gates how strongly it is applied by a per-batch coefficient that
interpolates between the uncorrected and the fully weighted gradient; the
coefficient combines an effective-sample-size term, a log-probability
divergence saturating at $\delta=0.02$, and a variance-amplification term
with tolerance $\gamma=1.2$ \citep{zhou_2026_ais}.

QuRL has two components that act at different levels \citep{li_2026_qurl}.
Adaptive Clipping Range (ACR) scales the PPO upper bound by the gap between
the quantized and the full-precision policy, $(1+\epsilon)/r_{i,t}$, which
presumes a learner left at full precision.  Our implementation follows
Equation~(9) of QuRL and does not apply dual-clip.  Because QuRL does not
specify the truncation cap $C$, we use $C=2$ in all QuRL runs.
Update-Aware Quantization (UAQ)
rescales the base model once before training, at the published scale $s=1.5$,
so that each RL update is large enough for the rollout quantizer to capture
rather than round away.  UAQ
needs the RMSNorm in front of a linear to absorb the rescale.  In an MoE
block that norm is shared with the router, and rescaling it would change
routing, so on Qwen3-30B-A3B we apply UAQ to the attention group only.
Table~\ref{tab:endpoint} evaluates this ACR+UAQ recipe, including the stated
MoE adaptation.  Because ACR is defined for a quantized rollout paired with
a full-precision learner, this is the applicable QuRL configuration;
substituting a QAT learner would change the policy pair that defines ACR.

Flash-RL uses verl's built-in rollout importance-sampling path.  No
public implementation of AIS or QuRL was available during our experiments;
we therefore reimplemented both methods from their papers and will release
our implementations with the code.

\subsection{Additional Design Analyses}
\label{app:design-analyses}

\paragraph{Support-aware allocation under a matched budget.}

\begin{figure}[!t]
  \centering
  \includegraphics[width=\linewidth]{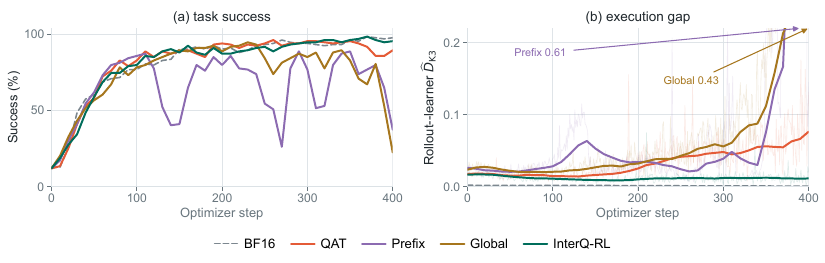}
  \caption{\captiona\ ALFWorld \texttt{valid\_unseen} success.
  \captionb\ Smoothed rollout--learner K3; arrows mark off-panel peaks.  All
  quantized arms share the W8A8-KV8+QAT substrate.}
  \label{fig:ablation}
\end{figure}

Figure~\ref{fig:ablation} provides the complete success and K3 trajectories
for the allocation variants analyzed in \S\ref{sec:exp-main}, together with
BF16 and QAT references; Prefix, Global, and InterQ-RL share the same bounded
prefix weights.
Over steps 301--400, median K3 is 0.0119 for InterQ-RL, 0.1172 for Prefix,
and 0.1858 for Global.  Appendix~B reports the allocated $\alpha_s$ ranges
over the full run.

\paragraph{Correction rules with learner alignment.}

\begin{table}[!t]
  \centering
  \caption{Token-local corrections with learner alignment on ALFWorld; all
  quantized arms use the W8A8-KV8 rollout and W/A+KV-QAT learner.}
  \label{tab:correction}
  \footnotesize
  \setlength{\tabcolsep}{2.0pt}
  \begin{tabular}{@{}lllcccc@{}}
  \toprule
  Variant & Scope & Retention & Success $\uparrow$ &
  AULC $\uparrow$ & Turns $\downarrow$ & Cap. $\downarrow$ \\
  \midrule
  BF16 & --- & --- & 97.14 & 82.07 & 10.98 & 2.24 \\
  \midrule
  QAT & none & none & 90.42 & \textbf{82.29} & 12.72 & 10.45 \\
  QAT+TIS & current token & soft clip $C=2$ & 90.55 & 70.59 & 9.86 & 5.22 \\
  QAT+AIS & current token & clip $C=5$ + gate & 92.66 & 76.27 & 12.46 & 6.72 \\
  InterQ-RL & segment prefix & $\alpha_s(\rho_s)$ & \textbf{96.39} & 81.96 & 10.06 & \textbf{4.48} \\
  \bottomrule
  \end{tabular}
  \par\vspace{2pt}
  {\raggedright Success averages the final six evaluations; Turns and Cap.
  use the step-400 evaluation.  AULC covers steps 0--400;
  Turns averages successful episodes and Cap. is the cap-exhaust
  percentage.\par}
\end{table}
As an additional ablation, the QAT+TIS and QAT+AIS rows pair Flash-RL's TIS
rule and the AIS correction rule, respectively, with the QAT learner used by
InterQ-RL.
With this learner configuration, InterQ-RL reaches 96.39\% last-six success, 3.73
points above the strongest token-local variant and 5.97 above QAT
(Table~\ref{tab:correction}).  Its AULC of 81.96 remains close to QAT's
82.29 and exceeds the token-local range of 70.59--76.27.  Successful
episodes average 10.06 turns, shorter than every variant except QAT+TIS, which
reaches 9.86 turns at only 90.55\% success.  InterQ-RL's 4.48\% cap-exhaust
rate is below the token-local range of 5.22\%--6.72\%.

\subsection{Implementation and hardware}
\label{app:hardware}

Training runs on verl \citep{sheng_2025_hybridflow} with vLLM
\citep{kwon_2023_vllm} as the rollout engine.  The 4B runs use the FSDP
backend; the 8B and 30B-A3B runs use the Megatron-LM backend with distributed
checkpointing.  All experiments run on NVIDIA H20 GPUs, with 8 GPUs per dense
run and 16 for 30B-A3B.

\clearpage
\raggedbottom
\section{Multi-Seed Replications and Budget Analyses}
\label{app:multiseed}

All tables in this appendix report $\text{mean} \pm \text{std}$ across three independent training seeds.

\subsection{Three-seed end-to-end comparison}
\label{app:multiseed-main}

\begin{table}[!ht]
\caption{Three-seed end-to-end comparison ($\text{mean} \pm \text{std}$, $n=3$) under a common training budget.
  ALFWorld and WebShop use Qwen3-4B; ScienceWorld headers give the backbone size.
  Bold marks the best quantized mean.}
\label{tab:multiseed}
\centering
\footnotesize
\setlength{\tabcolsep}{2.0pt}
\begin{tabular*}{\linewidth}{@{\extracolsep{\fill}}lccccccc@{}}
\toprule
& ALFWorld & \multicolumn{2}{c}{WebShop} &
\multicolumn{2}{c}{ScienceWorld (8B)} &
\multicolumn{2}{c}{ScienceWorld (A3B)} \\
\cmidrule(lr){2-2}\cmidrule(lr){3-4}\cmidrule(lr){5-6}\cmidrule(l){7-8}
Method & Success $\uparrow$ & Score $\uparrow$ & Success $\uparrow$ &
Macro $\uparrow$ & Solved $\uparrow$ & Macro $\uparrow$ & Solved $\uparrow$ \\
\midrule
BF16 & 93.3 $\pm$ 4.5 & 77.1 $\pm$ 0.1 & 53.5 $\pm$ 0.6 & 63.7 $\pm$ 2.8 & 72.9 $\pm$ 1.1 & 69.2 $\pm$ 4.0 & 75.5 $\pm$ 2.9 \\
\midrule
Flash-RL & 72.1 $\pm$ 28.2 & 74.6 $\pm$ 3.6 & 48.3 $\pm$ 6.8 & 62.3 $\pm$ 1.7 & 68.2 $\pm$ 1.2 & 66.6 $\pm$ 1.6 & 69.7 $\pm$ 1.7 \\
AIS & 90.0 $\pm$ 1.1 & 75.3 $\pm$ 3.0 & 51.1 $\pm$ 3.4 & 60.7 $\pm$ 1.3 & 63.6 $\pm$ 1.1 & 64.9 $\pm$ 1.0 & 70.5 $\pm$ 3.7 \\
QuRL & 77.1 $\pm$ 9.3$^\ddagger$ & 75.2 $\pm$ 0.7 & 50.9 $\pm$ 2.8 & 60.2 $\pm$ 3.2 & 64.9 $\pm$ 5.3 & 67.2 $\pm$ 2.8 & 72.0 $\pm$ 2.4 \\
QAT & 88.8 $\pm$ 2.7$^\dagger$ & 76.2 $\pm$ 1.2 & 51.7 $\pm$ 2.2 & 64.8 $\pm$ 1.6 & 72.6 $\pm$ 4.3 & 68.7 $\pm$ 0.6 & 74.8 $\pm$ 2.0 \\
InterQ-RL & \textbf{96.3} $\boldsymbol{\pm}$ \textbf{2.7} & \textbf{77.2} $\boldsymbol{\pm}$ \textbf{1.3} & \textbf{52.3} $\boldsymbol{\pm}$ \textbf{2.1} &
\textbf{66.1} $\boldsymbol{\pm}$ \textbf{1.6} & \textbf{74.7} $\boldsymbol{\pm}$ \textbf{1.3} & \textbf{69.9} $\boldsymbol{\pm}$ \textbf{1.0} & \textbf{77.1} $\boldsymbol{\pm}$ \textbf{1.2} \\
\bottomrule
\end{tabular*}
\par\vspace{2pt}
{\raggedright\footnotesize
$\dagger$ One of three QAT runs on ALFWorld collapsed after step 220 and is reported at its last pre-collapse evaluation (step 220, 85.8\%); the other two runs are reported at step 400.
$\ddagger$ All three QuRL runs on ALFWorld were unstable: two collapsed to 5--6\% success and are reported at their last pre-collapse evaluations (steps 200 and 300), and the third degraded late in training (reported at step 400).\par}
\end{table}

Table~\ref{tab:multiseed} reports step-400 endpoints across the four settings.
InterQ-RL gives the highest quantized mean on every metric, matching or exceeding BF16 on ALFWorld success (96.3\% versus 93.3\%), WebShop score (77.2 versus 77.1), and all four ScienceWorld metrics.

\begin{figure}[!ht]
  \centering
  \includegraphics[width=0.68\linewidth]{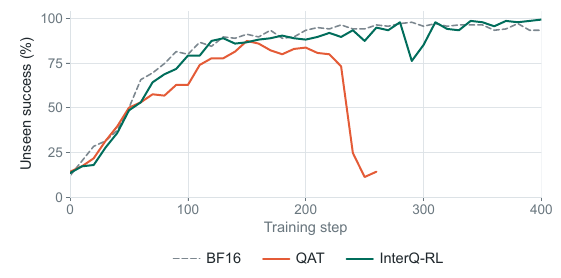}
  \caption{On ALFWorld (Qwen3-4B), another QAT seed reaches 87.3\% at step 150, close to BF16 and InterQ-RL, but collapses after step 220 to 14.2\% at step 260 while InterQ-RL remains stable through step 400.}
  \label{fig:qat-collapse}
\end{figure}

On ALFWorld, InterQ-RL improves over QAT by 7.5 points and over AIS by 6.3 points.
All three InterQ-RL runs remain stable through step 400.
For QAT, the run in Figure~\ref{fig:dynamics} rolls back from 95.5\% at step 300 to 89.6\% at step 400, and another seed collapses after step 220 to 14.2\% at step 260 (Figure~\ref{fig:qat-collapse}).
Table~\ref{tab:multiseed} credits collapsed QAT and QuRL runs with their last pre-collapse evaluation; using the collapsed QAT run's final score gives a three-seed mean of 64.9 $\pm$ 43.9\%.
On WebShop, InterQ-RL improves over QAT by 1.0 point on score and 0.6 on success; its score standard deviation (1.3) is less than half that of Flash-RL and AIS.

On ScienceWorld, InterQ-RL improves over QAT by 1.3 macro and 2.1 solved-rate points on Qwen3-8B and by 1.2 macro and 2.3 solved-rate points on Qwen3-30B-A3B.
On Qwen3-8B, InterQ-RL's solved-rate standard deviation (1.3\%) is less than one-third that of QAT (4.3\%).

\subsection{Three-seed allocation and budget analyses}
\label{app:multiseed-ablation}

\paragraph{Prefix-allocation variants across seeds.}
Table~\ref{tab:ablation-seeds} compares the allocation variants from Figure~\ref{fig:ablation}.
Prefix ($\alpha_s=1$) and Global ($\alpha_s=\alpha_g$) peak at 89.1\% and 94.5\% mid-training but fall to 51.7\% and 49.0\% by step 400.
Neither unallocated prefix weights nor uniform scaling under the same segment-normalized displacement budget sustains performance to step 400, whereas support-aware allocation reaches a 98.0\% peak and ends at 96.3\%.

\begin{table}[!ht]
  \centering
  \caption{Three-seed comparison ($\text{mean} \pm \text{std}$, $n=3$) of prefix-allocation variants on ALFWorld (Qwen3-4B); all quantized arms use the W8A8-KV8 rollout and W/A+KV-QAT learner.}
  \label{tab:ablation-seeds}
  \footnotesize
  \setlength{\tabcolsep}{5.0pt}
  \begin{tabular}{@{}llcc@{}}
  \toprule
  Variant & Allocation rule & Peak Success $\uparrow$ & Step 400 Success $\uparrow$ \\
  \midrule
  BF16 & --- & 97.5 $\pm$ 1.6 & 93.3 $\pm$ 4.5 \\
  \midrule
  QAT & none ($\alpha_s=0$) & 94.0 $\pm$ 6.0 & 88.8 $\pm$ 2.7$^\dagger$ \\
  Prefix & unallocated ($\alpha_s=1$) & 89.1 $\pm$ 1.1 & 51.7 $\pm$ 13.6 \\
  Global & uniform ($\alpha_s=\alpha_g$) & 94.5 $\pm$ 1.1 & 49.0 $\pm$ 23.1 \\
  InterQ-RL & support-aware ($\alpha_s(\rho_s)$) & \textbf{98.0} $\boldsymbol{\pm}$ \textbf{1.9} & \textbf{96.3} $\boldsymbol{\pm}$ \textbf{2.7} \\
  \bottomrule
  \end{tabular}
  \par\vspace{2pt}
  {\raggedright\footnotesize
  Peak Success is the maximum evaluation success over steps 0--400; $\dagger$ follows Table~\ref{tab:multiseed}.\par}
\end{table}

\paragraph{Sensitivity to the displacement budget.}
Table~\ref{tab:budget-ablation} scales the reference budget in \eqref{eq:alloc} as $\sum_{s\in\mathcal S} C_s\alpha_s^2 \le \kappa\,\alpha_g^2\sum_{s\in\mathcal S} C_s$ while retaining support-aware allocation by $\rho_s$.
Under both $\kappa=0.5$ and $\kappa=1.1$, all three seeds remain stable through step 400 without collapse, reaching 92.0\% and 91.3\% success and outperforming QAT (88.8\%). The default reference budget ($\kappa=1.0$) gives the highest step-400 success (96.3\%).

\begin{table}[!ht]
  \centering
  \caption{Three-seed sensitivity ($\text{mean} \pm \text{std}$, $n=3$) to the displacement-budget scale $\kappa$ in \eqref{eq:alloc} under support-aware allocation on ALFWorld (Qwen3-4B).}
  \label{tab:budget-ablation}
  \footnotesize
  \setlength{\tabcolsep}{5.0pt}
  \begin{tabular}{@{}llc@{}}
  \toprule
  Budget scale & Configuration & Step 400 Success $\uparrow$ \\
  \midrule
  $\kappa = 0.5$ & InterQ-RL (reduced budget) & 92.0 $\pm$ 1.6 \\
  $\kappa = 1.0$ & InterQ-RL (default budget) & \textbf{96.3} $\boldsymbol{\pm}$ \textbf{2.7} \\
  $\kappa = 1.1$ & InterQ-RL (increased budget) & 91.3 $\pm$ 2.3 \\
  \bottomrule
  \end{tabular}
\end{table}

\section{Runtime and Memory Profile}
\label{app:runtime}

Table~\ref{tab:runtime-profile} compares the runtime and memory profile of QAT and InterQ-RL on ALFWorld (Qwen3-4B, 400 steps, 128 trajectories per step from 16 prompts $\times$ 8 responses, $20{,}480$-token response budget).
Both arms share the vLLM W8A8-KV8 rollout engine and execute the same W/A+KV-QAT learner forward and backward passes (\S\ref{app:configs}).

\begin{table}[!ht]
  \centering
  \caption{Runtime and memory profile of QAT and InterQ-RL on ALFWorld
    (Qwen3-4B, 400 training steps).}
  \label{tab:runtime-profile}
  \footnotesize
  \setlength{\tabcolsep}{5.0pt}
  \begin{tabular*}{\linewidth}{@{\extracolsep{\fill}}lcc@{}}
  \toprule
  Metric & QAT & InterQ-RL \\
  \midrule
  Wall-clock time per update step (s) & 170.6 $\pm$ 32.1 & 171.8 $\pm$ 37.6 \\
  Total 400-step training time (h) & 18.96 & 19.09 \\
  Prefix statistics \& allocation solver (Algorithm~\ref{alg:interq}, ms/step) & --- & 31.3 \\
  Peak GPU memory, allocated / reserved (GB) & 17.37 / 32.72 & 17.39 / 32.70 \\
  \bottomrule
  \end{tabular*}
  \par\vspace{2pt}
  {\raggedright\footnotesize
  Per-step metrics are $\text{mean} \pm \text{std}$ across the 400 training steps ($3{,}072$ action segments per step).\par}
\end{table}

On the learner side, Algorithm~\ref{alg:interq} reuses the existing learner and rollout log probabilities $(\log q_{\mathrm{old}}, \log r_{\mathrm{exec}})$ and takes $31.3\text{ ms}$ per update step ($0.018\%$ of step time).
Masked prefix accumulation takes $18.6\text{ ms}$ (lines 1--4), segment extraction and statistics $8.9\text{ ms}$ (lines 5--7), the 80-step bisection solver $1.5\text{ ms}$ (lines 8--9), and weight assembly $2.3\text{ ms}$ (line 10).
Over 400 steps, end-to-end wall-clock time differs by $0.7\%$ ($19.09\text{ h}$ for InterQ-RL versus $18.96\text{ h}$ for QAT), and peak allocated and reserved GPU memory is essentially unchanged ($17.39 / 32.70\text{ GB}$ versus $17.37 / 32.72\text{ GB}$).
All quantized arms share the same W8A8-KV8 decoding path, which InterQ-RL does not modify, so rollout throughput and KV-cache savings are identical.

\end{document}